\pdfoutput=1

\documentclass[10pt,twocolumn,letterpaper]{article}

\usepackage{times}
\usepackage{epsfig}
\usepackage{graphicx}
\usepackage{amsmath}
\usepackage{amssymb}
\usepackage{fancyhdr}
\usepackage{fixltx2e}
\usepackage{subcaption}
\usepackage{ulem}

\usepackage[pagebackref=true,colorlinks,bookmarks=false]{hyperref}

\usepackage[ruled,vlined,linesnumbered]{algorithm2e}
\usepackage{algpseudocode}

\usepackage{booktabs}
\usepackage{tablefootnote}
\usepackage{array}

\newcommand{\eat}[1]{}

\usepackage{amsmath}
\usepackage{amssymb}
\usepackage{amsthm}

\newcommand{\myvec}[1]{\mathbf{#1}}
\newcommand{\myvecsym}[1]{\boldsymbol{#1}}

\newcommand{\vtheta}{\myvecsym{\theta}}

\newcommand{\vx}{\myvec{x}}

\newcommand{\vy}{\myvec{y}}

\newcommand{\mymathcal}[1]{\mathcal{#1}}

\newcommand{\calD}{{\mymathcal{D}}}

\newcommand{\calG}{\mymathcal{G}}

\newcommand{\calV}{\mymathcal{V}}

\newcommand{\be}{\begin{equation}}
\newcommand{\ee}{\end{equation}}
\newcommand{\bea}{\begin{eqnarray}}
\newcommand{\eea}{\end{eqnarray}}
\newcommand{\beaa}{\begin{eqnarray*}}
\newcommand{\eeaa}{\end{eqnarray*}}
\newcommand{\ba}{\begin{align*}}
\newcommand{\ea}{\end{align*}}

\newenvironment{compactitemize}
{ \begin{itemize}
    \addtolength{\itemindent}{-0.4cm}
    \setlength{\itemsep}{0pt}
    \setlength{\parskip}{0pt}
    \setlength{\parsep}{0pt}     }
{ \end{itemize}                  } 

\newenvironment{compactenumerate}
{ \begin{enumerate}
    \addtolength{\itemindent}{-0.4cm}
    \setlength{\itemsep}{0pt}
    \setlength{\parskip}{0pt}
    \setlength{\parsep}{0pt}     }
{ \end{enumerate}                  }

\usepackage{caption} 
\usepackage{authblk}
\usepackage[none]{hyphenat}

\begin{document}


\title{Improved Image Captioning via Policy Gradient optimization of SPIDEr}

\author[1]{Siqi Liu\thanks{The major part of this work was done while Siqi Liu was an intern at Google.}}
\author[2]{Zhenhai Zhu}
\author[2]{Ning Ye}
\author[2]{Sergio Guadarrama}
\author[2]{Kevin Murphy}
\affil[ ]{\texttt {siqi.liu@cs.ox.ac.uk}}
\affil[ ]{\texttt {\{zhenhai,nye,sguada,kpmurphy\}@google.com}}
\affil[1]{Department of Computer Science, University of Oxford}
\affil[2]{Google}

\date{}

\maketitle

\begin{abstract} 
	
	Current image captioning methods are usually trained via (penalized)
	maximum likelihood estimation. However, the log-likelihood score of a
	caption does not correlate well with human assessments of quality.
	Standard syntactic evaluation metrics, such as BLEU, METEOR and ROUGE,
	are also not well correlated.  The newer SPICE and CIDEr metrics are
	better correlated, but have traditionally been hard to optimize for.  In
	this paper,  we show how to use a policy gradient (PG) method to
	directly optimize a linear combination of SPICE and CIDEr (a combination
	we call SPIDEr): the SPICE score ensures our captions are semantically
	faithful to the image, while CIDEr score ensures our captions are
	syntactically fluent. The PG method we propose improves on the prior
	MIXER approach, by using Monte Carlo rollouts instead of mixing MLE
	training with PG.  We show empirically that our algorithm leads to easier
	optimization and improved results compared to MIXER. Finally, we show
	that using our PG method we can optimize any of the metrics, including
	the proposed SPIDEr metric which results in image captions that are
	strongly preferred by human raters compared to captions generated by the
	same model but trained to optimize MLE or the COCO metrics.

\end{abstract}

\section{Introduction}
\label{sec:intro}

Image captioning is the task of describing the visual content of an image using
one or more sentences.  This has many applications,
including text-based image retrieval, accessibility for blind users \cite{Wu2017}, and
human-robot interaction \cite{das2016visual}.

Most methods for solving this task require training a statistical model
on a dataset of (image, caption) pairs.
The model is usually trained to maximize the log likelihood of the training set.
\eat{
which is given by

\begin{align*}
L(\theta) 
&= \frac{1}{N} \sum_{n=1}^N \log p(\vy^n|\vx^n,\vtheta)\\
&= \frac{1}{N} \sum_{n=1}^N \sum_{t=1}^{T_n} \log p(y_t^n|y^n_{1:t-1},\vx^n,\vtheta)
\end{align*}

where $\vx^n$ is the $n$'th image, $\vy^n=(y^n_1,\ldots,y^n_{T_n})$
is the ground truth caption of the $n$'th image and $N$ is the total number of
labelled examples.
}
After training, these models are usually evaluated by computing
a variety of different metrics on a test set,
such as COCO \cite{Lin2014coco}.
Standard metrics from the machine translation community include
BLEU \cite{Papineni2002bleu},
METEOR \cite{Banerjee2005meteor},
and ROUGE \cite{Lin2003rouge}.
More recently, the
CIDEr metric \cite{Vedantam2015} was proposed, specifically for the
image captioning task.
We shall call the combination of these four metrics ``BCMR'', for short.
Unfortunately, none of these metrics correlate strongly with human measures of caption quality.
In fact, humans score lower on these metrics than the methods that won the COCO 2015 challenge,
despite the fact that humans are still much better at this task.

These results motivated Anderson et al.
\cite{Anderson2016} to propose the SPICE metric.
Rather than directly comparing a generated sentence to a set of reference sentences
in terms of syntactic agreement,
SPICE  first 
parses each of the reference sentences, and then uses them
to derive an abstract scene graph representation.
The generated sentence is then also parsed, and compared to the graph;
this allows for a comparison of the semantic similarity,
without paying attention to syntactic factors
(modulo the requirement that the generated sentence be parseable).
\cite{Anderson2016} showed that SPICE is the only existing metric that has a
strong correlation with human ratings, and ranks human captions above
algorithms submitted to the COCO benchmark.

Given this result, it is natural to want to directly optimize SPICE.
However, this is tricky, since it is not a differentiable objective.
In this paper, we show that it is possible to use policy gradient (PG)
methods \cite{Sutton1999} to optimize such objectives.
The idea of using PG to optimize non differentiable objectives for
image captioning was first proposed in the MIXER paper
\cite{Ranzato2015}. However, they only used it to optimize BLEU-4,
which is not correlated with human quality.
Furthermore, when we tried to use their method to optimize
other metrics, such as CIDEr or SPICE, we
got poor results, since their method
(which involves an intricate incremental schedule which mixes from MLE
training to full PG training) is not very robust and require careful tunning.

In this paper, we  propose an improvement to MIXER,
that uses Monte Carlo rollouts to get a better estimate of the value function,
and avoids the need to ``mix in'' the MLE objective.
We show that this leads to faster convergence, and is significantly
more robust to choice of learning rates and other hyper-parameters.
We then show that we can use our new PG method
to optimize the BCMR metrics, thus achieving state of the art results
on the COCO leaderboard, despite using a very simple baseline model.

However, being on top of the COCO leaderboard is not our ultimate
goal, since we know that the COCO metrics (based on BCMR)
are only weakly correlated with human judgement \cite{Anderson2016}.
So we decided to optimize SPICE with our algorithm.
Unfortunately, optimizing SPICE produced long repetitive sentences which
are not good results, based either on BMCR COCO metrics, or as judged by humans.

The reason optimizing SPICE gives poor captions
is that SPICE ignores syntactic quality (see examples in Table \ref{fig:examples})
More generally, we argue that
a good image captioning metric should satisfy two criteria:
  (1) captions that are considered good by humans should achieve high
  scores; and
  (2) captions that achieve high scores should be  considered good by
  humans.
  SPICE satisfies criterion 1, but not criterion 2.
  We therefore propose a new metric, which is a linear combination of SPICE and
  CIDEr;  we call this new metric SPIDEr.
  This metric automatically satisfies criterion 1, since both SPICE and CIDEr do.
  Also, when we optimize for SPIDEr, we show that the captions are judged by
humans to be significantly better than the ones generated by training
the same model using any of the other metrics.
This shows that SPIDEr satisfies criterion 2 to a much greater degree than existing metrics.

In summary, we make the following contributions in this paper:
(1) we identify criteria for a good image captioning metric,
and proposed a new metric, SPIDEr, which meets both criteria;
(2) we propose a new policy gradient method that can optimize
arbitrary captioning metrics, and which is much faster and more stable
than previous PG methods;
(3) we show that using our new PG method to optimize
existing BCMR metrics leads to
state of the art results on COCO;
(4) we show that using our new PG method to optimize our new SPIDEr metric
results in much better human scores than optimizing for other metrics.

\section{Related work}
\label{sec:related}

There is an extensive body of work on image captioning
(see e.g., \cite{Bernardi2016} for a recent review).
Below we summarize some of the most relevant works.

\subsection{Models}

Most methods make use an encoder-decoder style neural network,
where the encoder is a convolutional neural network (CNN),
and the decoder a recurrent neural network (RNN).
In this work, we use the encoder-decoder proposed
in \cite{Vinyals2015}, known as ``Show and Tell'' (ST).

Numerous extensions to the basic encoder-decoder framework have been
proposed.
One line of work (e.g., the ``Show, Attend and Tell'' model of
\cite{Xu2015},
and the ``Review Network'' model of \cite{Yang2016}),
leverages attention, which lets  the decoder focus on specific parts
of the input image when generating a word.
Another line of work 
enriches the image encoding beyond just using a CNN that was trained
for image classification. For example,
 \cite{Wu2016,Yao2016} use image taggers,
 \cite{Venugopalan2016} use object detectors,
and \cite{Tran2016} use
face detection and landmark recognition.
We stress that these extensions are orthogonal to the ideas in this
paper.

\subsection{Metrics and objective functions}

Most prior work uses maximum likelihood estimation (MLE) for training. That is,
the model parameters $\vtheta$ are trained to maximize \vspace{-15pt}

\begin{align*}
L(\theta) 
&= \frac{1}{N} \sum_{n=1}^N \log p(\vy^n|\vx^n,\vtheta)\\
&= \frac{1}{N} \sum_{n=1}^N \sum_{t=1}^{T_n} \log p(y_t^n|y^n_{1:t-1},\vx^n,\vtheta)
\end{align*}

where $\vx^n$ is the $n$'th image, $\vy^n=(y^n_1,\ldots,y^n_{T_n})$
is the ground truth caption of the $n$'th image and $N$ is the total number of
labelled examples.

One problem with the MLE objective is that
at training time, each prediction is conditioned on the
previously observed words from the ground truth. At test time, however, the
model will be fed with its own predictions, leading to quickly accumulating
errors during inference, so the model will likely diverge
from desired trajectories. This discrepancy is known as ``exposure bias''
\cite{Bengio2015}.
One solution to exposure bias is  ``scheduled
sampling'' \cite{Bengio2015},
although this method has been shown to be statistically
inconsistent \cite{Huszar2015}.

An alternative to maximizing likelihood is to try to maximize some
other objective that is more closely related to the true metric of
interest.
This can be done using a policy gradient (PG)
method \cite{Sutton1999} such as REINFORCE \cite{Williams1992},
by treating the score of a candidate
sentence as analogous to a reward signal in a reinforcement learning setting.
In such a framework, the RNN decoder acts like a stochastic policy, where
choosing an action corresponds to generating the next word.

\cite{Ranzato2015} use 
a modified form of REINFORCE
to optimize the BLEU score for image captioning.
We explain this approach in more detail in Section~\ref{sec:mixer},
since it is closely related to our method.
\cite{hendricks2016generating}
used  REINFORCE to optimize sequence level reward such that generated captions are
class discriminative. However, this is a different task than the one
we consider.

More recently, there have been a variety of other papers on
optimizing sequence level objective functions
(see
e.g., \cite{Norouzi2016,Yu2016,Bahdanau2016,Shen2016}),
but mostly in the context of machine
translation. 

\subsection{MIXER}
\label{sec:mixer}

The most closely related work is the ``MIXER'' paper \cite{Ranzato2015}.  They
also use the REINFORCE method, combined with a baseline reward estimator.
However, they implicitly assume each intermediate action (word) in a partial
sequence has the same reward as the sequence-level reward, which is not true in
general.  To compensate for this, they introduce a form of training that mixes
together the MLE objective and the REINFORCE objective.  Specifically, they
evaluate the first $M$ words using log-likelihood, and the remaining words using
REINFORCE; they gradually decrease $M$ from the maximum sentence length down to
0.

We have found that MIXER is very sensitive to the form of the annealing
schedule, as well as other hyper-parameters, such as the learning rate or
gradient scaling of the baseline estimator. We therefore propose a more robust
alternative, in which we replace the constant reward assumption with an estimate
of future rewards based on Monte Carlo rollouts (a similar idea is used in
\cite{Yu2016} for GAN training).  In Section~\ref{sec:results}, we show that
this change significantly improves convergence speed, as well as training
stability.  This lets us easily optimize a variety of different metrics,
including our new metric, SPIDEr.

\section{Methods}
\label{sec:methods}

In this section, we explain our approach in more detail.  First we discuss the
policy gradient method, which can be used to robustly optimize any kind of
reward function. Next we discuss which reward function to use. Finally, we
discuss the model itself, which is a standard CNN-RNN.

\subsection{Training using policy gradient}
\label{sec:PG}

\begin{figure}
    \centering
    \includegraphics[width=0.7\linewidth]{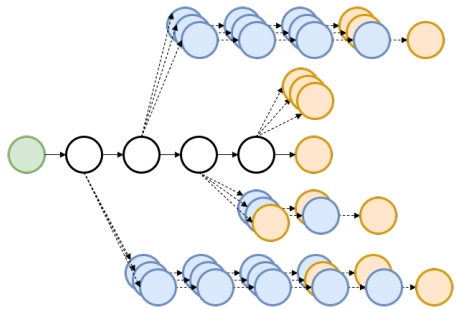}
    \caption{The value of each action is estimated as the average rewards
      received by its $K$ rollout sequences (i.e. $K=3$). Solid arrows indicate the
      sequence of actions being evaluated. The tokens in green and yellow are
      respectively \texttt{BOS} (beginning of sequence) and \texttt{EOS}
      (end of sequence) tokens. Sequences in blue are rollout sequences sampled
      from partial sequences. Note that rollout sequences do not always have the
      same length, as they are separately sampled from a stochastic policy.}
  \label{fig:mc_rollouts} \end{figure}

At time step $t$, we pick a discrete action,
which corresponds to choosing a word $g_t \in \calV$,
using a stochastic policy or generator $\pi_{\theta}(g_t|s_t,\vx)$,
where $s_t=g_{1:t-1}$ is the sequence of words chosen so far, $\vx$ is the 
image and $\theta$ are the parameters of the model.
Note that in our case the state transition function is deterministic: we simply
append the word chosen at time $t$ to get $s_{t+1}=s_t; g_t$
($u;v$ is the concatenation of the strings $u$ and $v$).

When we reach the end of the sequence (i.e., once the generator
emits the end-of-sentence marker), we get a reward of $R(g_{1:T}|\vx^n,\vy^n)$,
which is  the score  for producing caption $g_{1:T}$ given image $\vx^n$
and ground truth caption (or set of captions) $\vy^n$.
This reward can be any function, such as BCMR or SPICE.

In typical reinforcement learning setting, an agent receives rewards at each
intermediate step while future rewards are discounted to balance short-term and
long-term gain. In our case, however, the agent receive zero reward
during intermediate steps, observing a reward only at the end. To mitigate the
lack of intermediate reward signal, we propose to estimate the value of
intermediate states (partial sequence), via Monte-Carlo rollouts. This translate
into significantly more robust credit assignment and efficient gradient
estimation, as we show in Section~\ref{sec:results}. We define the value
function of a partial sequence as its expected future reward:

\be
V_{\theta}(g_{1:t}|\vx^n,\vy^n) =
E_{g_{t+1:T}} [R(g_{1:t}; g_{t+1:T}|\vx^n, \vy^n)]
\ee
where the expectation is w.r.t.
$g_{t+1:T} \sim \pi_{\theta}(\cdot| g_{1:t}, \vx^n)$.

Our goal is to maximize the average reward starting
from the initial (empty) state $s_0$ defined as:
\be
J(\theta) = \frac{1}{N} \sum_{n=1}^N V_{\theta}(s_0 | \vx^n, \vy^n)
\label{J_theta}
\ee
where $N$ is the number of examples in the training set.
We now discuss how to optimize Eqn. (\ref{J_theta}).
For simplicity, we will consider a single example $n$,
so 
we will drop the $\vx^n$ and $\vy^n$ notation.
To compute the gradient of $J(\theta)$, we can use
the policy gradient theorem from
\cite{Sutton1999}.
In the special case of deterministic transition functions,
this theorem simplifies as shown below
(see \cite{Bahdanau2016}
for a proof):
\be
\nabla_{\theta} V_{\theta}(s_0)
 = E_{g_{1:T}} \left[ \sum_{t=1}^T \sum_{g_t \in \calV}
  \nabla_{\theta} \pi_{\theta}(g_t | g_{1:t-1})  Q_{\theta}(g_{1:t-1}, g_t) \right]
\ee
where we define the $Q$ function
for a state-action pair as follows:
\be
Q_{\theta}(g_{1:t-1}, g_t) = E_{g_{t+1:T}} [R(g_{1:t-1}; g_t; g_{t+1:T})]
\ee
We can approximate the gradient of the value function with
$M$  sample paths, $g_{1:T}^m \sim \pi_{\theta}$, generated from our policy.
This gives 
\begin{align}
\nabla_{\theta} V_{\theta}(s_0)
 &\approx \frac{1}{M} \sum_{m=1}^M \sum_{t=1}^T E_{g_t} \left[
  \nabla_{\theta} \log \pi_{\theta}(g_t | g_{1:t-1}^m)
  \right. \nonumber \\
  & \left.
  \times  Q_{\theta}(g_{1:t-1}^m, g_t) \right]
\label{eqn:deriv}
\end{align}
where the expectation is w.r.t.
$g_t \sim \pi_{\theta}(g_t|g_{1:t-1}^m)$,
and where
we have exploited the fact that
\[
\nabla_{\theta} \pi_{\theta}(a|s) =
\pi_{\theta}(a|s) \frac{\nabla_{\theta} \pi_{\theta}(a|s)}
   {\pi_{\theta}(a|s)}
   = \pi_{\theta}(a|s) \nabla_{\theta} \log \pi_{\theta}(a|s)
\]
If we use $M=1$,
we can additionally replace the $E_{g_t}$ with the value in the sample
path, $g_t^m$, as in REINFORCE. In our experiment, we used $M=1$ and we
subsequently drop the superscript $m$ in the rest of this paper for notational
clarity.
   
The only remaining question is how to estimate
the function $Q(s_t,g_t)$.
For this, we will follow \cite{Yu2016} and use Monte Carlo rollouts.
In particular,
we first sample $K$ continuations of the sequence
$s_t;g_t$ to get $g_{t+1:T}^k$. Then we compute the average
\be
Q(g_{1:t-1},g_t) \approx \frac{1}{K} \sum_{k=1}^K R(g_{1:t-1}; g_t; g_{t+1:T}^k)
\label{eqn:rollout}
\ee
We estimate how good a 
particular word choice $g_t$ is by averaging over all complete sequences sampled
according to the current policy, conditioned on the partial sequence 
$g_{1:t-1}$ sampled from the current policy so far.
This process is illustrated in Figure~\ref{fig:mc_rollouts}.
If we are in a terminal state,
we define $Q(g_{1:T}, \mathtt{EOS}) = R(g_{1:T})$. 

The above gradient estimator is an unbiased but high variance estimator. 
One way to reduce its variance is to estimate the expected baseline reward 
$E_{g_t}[Q(g_{1:t-1}, g_t)]$ using a parametric function;
we will denote this baseline as 
$B_{\phi}(g_{1:t-1})$.
We then subtract this baseline from
$Q_{\theta}(g_{1:t-1}, g_t)$ 
to get the following estimate for the gradient
(using $M=1$ sample paths):
\begin{align}
\nabla_{\theta} V_{\theta}(s_0)
&\approx  \sum_{t=1}^T \sum_{g_t} \left[
  \pi_{\theta}(g_t|s_t)
\nabla_{\theta} \log \pi_{\theta}(g_t | s_t) \right. \nonumber \\
& \left. \times
  \left(  Q_{\theta}(s_t, g_t)  - B_{\phi}(s_t) \right) \right]
\label{eqn:gradient}
\end{align}
where $s_t = g_{1:t-1}$.
Subtracting the baseline does not affect the validity of the estimated gradient,
but reduces its variance. Here, we simply refer to prior work 
(\cite{zaremba2015reinforcement}, \cite{Williams1992}) for a full derivation of 
this property. 

We train the parameters $\phi$ of the baseline estimator
to minimize the following loss:
\be
L_{\phi} = \sum_t E_{s_t} E_{g_t} (Q_{\theta}(s_t, g_t) - B_{\phi}(s_t))^2
\ee
In our experiments, the baseline estimator is an MLP which takes as input the
hidden state of the RNN at step $t$. To avoid creating a feedback loop, we do 
not back-propagate gradients through the hidden state from this loss.

In language generation settings, a major challenge facing PG methods is the
large action space. This is the case in our task, where the action space
corresponds to the entire vocabulary of 8,855 symbols. To help ``warm start''
the training, we pre-train the RNN decoder (stochastic policy) using
MLE training, before switching to PG training. This prevents the agent
from performing random walks through exponentially many possible paths at the
beginning of the training.

The overall algorithm is summarized in
Algorithm~\ref{algo:PG}.
Note that the Monte Carlo rollouts only require a forward pass through the
RNN, which is much more efficient than the forward-backward pass needed for the
CNN.  Additionally the rollouts can be also be done in parallel for multiple
sentences.  Consequently, PG training is only about twice as slow as MLE
training (in wall time).

\begin{algorithm}
\caption{PG training algorithm}
\label{algo:PG}
Input: $\calD = \{ (\vx^n, \vy^n): n=1:N \}$ \;
Train $\pi_{\theta}(g_{1:T}|x)$ using MLE on $\calD$ \;
Train $B_{\phi}$ using MC estimates of $Q_{\theta}$ on a small subset of $\calD$\;
\For{each epoch}{
  \For{example $(x^n, y^n)$}{
    Generate sequence $g_{1:T} \sim \pi_{\theta}(\cdot|x^n)$ \;
    \For{$t=1:T$}{
      Compute $Q(g_{1:t-1}, g_t)$ for $g_t$ with $K$ Monte Carlo rollouts, 
        using (\ref{eqn:rollout})\;
      Compute estimated baseline $B_{\phi}(g_{1:t-1})$\;
    }
    Compute $\calG_{\theta} = \nabla_{\theta} V_{\theta}(s_0)$ using
    (\ref{eqn:gradient})\;
    Compute $\calG_{\phi} = \nabla_{\phi} L_{\phi}$\;
    SGD update of $\theta$ using $\calG_{\theta}$\;
    SGD update of $\phi$ using $\calG_{\phi}$\;
  }
}
\end{algorithm}

\subsection{Reward functions for the policy gradient}

We can use our PG method to optimize many different reward functions.
Common choices include
BLEU, CIDEr, METEOR and ROUGE.
Code for all of these metrics is available as part of the COCO evaluation toolkit.\footnote{
  \url{https://github.com/tylin/coco-caption}.
  }
We decided to use a weighted combination of all of these.
Since these  metrics are not on the same scale, we chose in our experiments
the set of weights such that all metrics have approximately the same magnitude.
More precisely, we choose the following weighted combination:
0.5*BLEU-1 +  0.5*BLEU-2 + 1.0*BLEU-3 + 1.0*BLEU-4 + 1.0*CIDEr + 5.0*METEOR +  2.0*ROUGE.
Optimizing this weighted combination of BCMR gives state-of-the-art results on
the COCO test set, as we discuss in Section~\ref{sec:coco}.

One problem with the BCMR metrics is that individually they are not well correlated
with human judgment \cite{Anderson2016}. We therefore also tried optimizing the
recently introduced SPICE metric \cite{Anderson2016}, which better reflects human estimates of quality.
We use the open source release of the SPICE code\footnote{
  \url{https://github.com/peteanderson80/SPICE}.
}
to evaluate the metric.

Interestingly, we have found that just optimizing SPICE tended to result in
captions which are very detailed, but which
often had many repeated phrases,
as we show in Section~\ref{sec:results}.
This is because SPICE 
measures semantic similarity  (in terms of a scene graph) between sets
of sentences, but does not pay attention to syntactical factors
(modulo the requirement that the generated sentence be parseable).
We therefore combined SPICE with the CIDEr metric (considered the best
of the standard automatic metrics for COCO), a combination we
call SPIDEr for short.
Based on initial experiments, we decided to use an equal weighting for both.

\subsection{Encoder-decoder architecture}
\label{sec:model}

\begin{figure}
    \centering
    \includegraphics[width=0.8\linewidth]{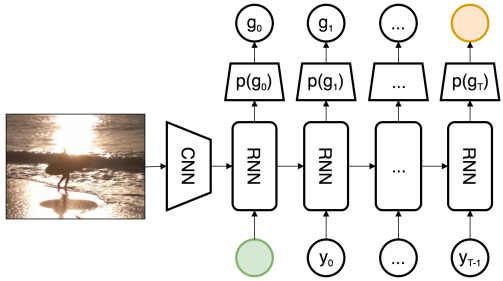}
    \caption{Model architecture of Show and Tell image captioning
      system~\cite{Vinyals2015}.  The tokens in green and yellow are
      respectively \texttt{BOS} (beginning of sequence) and \texttt{EOS}
      (end of sequence) tokens. At testing time, output from previous time
      step $g_{t-1}$ is used as input in lieu of $y_{t-1}$.}
    \label{fig:show_tell}
\end{figure}

We use a CNN-RNN architecture similar to the one proposed in the original
Show-Tell paper \cite{Vinyals2015}. A high-level diagram is shown in
Figure~\ref{fig:show_tell}. Each symbol in the vocabulary is embedded as
a 512 dimensional dense word embedding vector, whose values are initialized 
randomly.

The encoder CNN is implemented as an Inception-V3~\cite{szegedy2015rethinking}
network pretrained on ImageNet\footnote{We used the open-source implementation
available at:
\url{https://github.com/tensorflow/models/blob/master/slim/nets/inception_v3.py}}.The
RNN decoder is a one-layer LSTM with a state size of $512$ units, initialized
randomly. Each image is encoded by Inception-V3 as a dense feature vector of
dimension $2,048$ which is then projected to $512$ dimension with a linear layer
and used as the initial state of RNN decoder.

At training time, we always feed in the ground truth symbol to the RNN decoder;
at inference time we use just greedy decoding, where the sampled output is fed to the
RNN as the next input symbol.
\eat{
We use a greedy decoding procedure, in which we pick the most probable
word at each time step. Note that this is suboptimal, since
\be
\prod_{t=1}^T \max_{g_t} p(g_t|g_{1:t-1})
\leq \max_{g_{1:T}} \prod_{t=1}^T p(g_t|g_{1:t-1})
\ee
We can better approximate the globally optimal sequence by using beam search.
However, we find that models trained with PG do not benefit from this in
practice, since they tend to concentrate their probability mass on a very small
number of actions at each step.
}

\section{Results}
\label{sec:results}

\subsection{Experimental protocol}

We report results obtained by different methods on the COCO
dataset.  This has 82,081 training images, and 40,137 validation images, each
with at least 5 ground truth captions. Following standard practice for methods
that evaluate on the COCO test server,  we hold out a small subset of 1,665
validation images for hyper-parameter tuning, and use the remaining combined
training and validation set for training.

We preprocess the text data by lower casing, and replacing words which occur
less than 4 times in the 82k training set with UNK; this results in a vocabulary
size of 8,855  (identical to the one used in \cite{Vinyals2015}).  At training
time, we keep all captions to their maximum lengths. At testing time, the
generated sequences are truncated to 30 symbols in all experiments.

We use the Show-Tell model from \cite{Vinyals2015} for all methods.
We ``pre-train'' this model with MLE, and then optionally ``fine tune'' it
with other methods, as we discuss below.

\subsection{Automatic evaluation using BCMR metrics}
\label{sec:coco}

\begin{table*}[ht]
\centering
\begin{tabular}{@{}lccccccc@{}}
\toprule
Submissions & \multicolumn{1}{l}{CIDEr-D} & \multicolumn{1}{l}{Meteor} & \multicolumn{1}{l}{ROUGE-L} & \multicolumn{1}{l}{BLEU-1} & \multicolumn{1}{l}{BLEU-2} & \multicolumn{1}{l}{BLEU-3} & \multicolumn{1}{l}{BLEU-4} \\ \midrule
MSM@MSRA~\cite{Yao2016} & 0.984 & 0.256 & 0.542 & 0.739 & 0.575 & 0.436 & 0.330 \\
Review Net~\cite{Yang2016} & 0.965 & 0.256 & 0.533 & 0.720 & 0.550 & 0.414 & 0.313 \\
ATT~\cite{You2016} & 0.943 & 0.250 & 0.535 & 0.731 & 0.565 & 0.424 & 0.316 \\
Google~\cite{Vinyals2015} & 0.943 & 0.254 & 0.530 & 0.713 & 0.542 & 0.407 & 0.309 \\
Berkeley LRCN~\cite{donahue2015long} & 0.921 & 0.247 & 0.528 & 0.718 & 0.548 & 0.409 & 0.306 \\
\midrule
MLE & 0.947 & 0.251 & 0.531 & 0.724 & 0.552 & 0.405 & 0.294 \\
PG-BLEU-4 & 0.966 & 0.249 & 0.550 & 0.737 & 0.587 & \textbf{0.455} & \textbf{0.346} \\
PG-CIDEr & 0.995 & 0.249 & 0.548 & 0.737 & 0.581 & 0.442 & 0.333 \\
\midrule
MIXER-BCMR & 0.924 & 0.245 & 0.532 & 0.729 & 0.559 & 0.415 & 0.306 \\
MIXER-BCMR-A & 0.991 & \textbf{0.258} & 0.545 & 0.747 & 0.579 & 0.431 & 0.317 \\
PG-BCMR & \textbf{1.013} & \textbf{0.257} & \textbf{0.55} & \textbf{0.754} & \textbf{0.591} & 0.445 & 0.332 \\ 
PG-SPIDEr & 1.000 & 0.251 & 0.544 & 0.743 & 0.578 & 0.433 & 0.322 \\
\end{tabular}
\caption{Automatic evaluation on the official COCO C-5 test split,
  as of November 2016.
Methods below the line are contributions from this paper.
}
\label{fig:leaderboard}
\end{table*}

In this section, we quantitatively evaluate various methods using the standard
BCMR metrics on the COCO test set.  Since the ground truth is not available for
the test set, we submitted our results to the COCO online evaluation
server\footnote{\url{mscoco.org/dataset/\#captions-leaderboard}} on Nov 2016.
Table~\ref{fig:leaderboard} shows the results of the top 5 methods (at the time of
submission) on the official C-5 leaderboard, along with the results
of our experiments.  In particular, we tried training the Show-Tell model with
the following methods: MLE, PG-BLEU-4, PG-CIDEr, PG-BCMR, PG-SPIDEr (with equal
weight on SPICE and on CIDEr), and MIXER.

Our MLE method gets similar results to the scheduled sampling method of
\cite{Vinyals2015}. Not surprisingly, PG-BCMR  significantly outperforms MLE
training, since it directly optimizes for BCMR.  Similarly, PG-BCMR outperforms
PG-SPIDEr. We also showed that PG-BLEU-4 and PG-CIDEr specifically improves on
the metric optimized for, compared to MLE baseline. In particular, PG-BLEU-4
successfully achieved the highest score on BLEU metrics, while largely neglected
others. This demonstrates the general applicability of our optimization method
to target any specific metric of interests. However, as shown below optimizing
for the current COCO metrics does not translate into better captions.

We also see that our PG-BCMR method significantly outperforms all the top 5
methods, even the ones which use more sophisticated models, such as those based
on attention (Montreal/Toronto, ATT, Review Net), those that use more complex
decoders (Berkeley LRCN), and those that use high-level visual attributes
(MSM@MSRA, ATT).

Our PG-BCMR method also outperforms the MIXER algorithm;
we discuss this in more detail in Section~\ref{sec:mixerResults}.

\subsection{Human evaluation}
\label{sec:human}


\begin{table*}[tbp]
  \centering
  \scriptsize
  \begin{tabular}{ | c | m{5cm} | m{6cm} | }
    \hline
    Images & Ground Truth Captions & Generated Captions \\ \hline

    \begin{minipage}{.2\textwidth}
      \includegraphics[width=\linewidth]{Figures/COCO_val2014_000000057363.jpg}
    \end{minipage}
    &
      \begin{compactenumerate}
        \item a red and yellow fire truck and some buildings
        \item An overhead view shows a fire engine in the street.
        \item A red and yellow fire truck with ladders on top
        \item A firetruck is parked in the street in between stop lights.
        \item A fire truck (ladder truck) drives down a street in the city.
      \end{compactenumerate}
    &
      \begin{compactitemize}
        \item MLE: a red and white bus is driving down the street
        \item PG-SPICE: a red double decker bus on a city street on a street with a bus on the street with a bus on the street in front of a bus on
        \item MIXER-BCMR: a yellow bus driving down a city street .
        \item MIXER-BCMR-A: a red fire truck driving down a city street .
        \item PG-BCMR: a red bus driving down a city street .
        \item PG-SPIDEr: a red fire truck is on a city street.
      \end{compactitemize}
    \\ \hline

    \begin{minipage}{.2\textwidth}
      \includegraphics[width=\linewidth]{Figures/COCO_val2014_000000541924.jpg}
    \end{minipage}
    &
      \begin{compactenumerate}
        \item A woman walking on a city street in a red coat.
	\item A group of people that are standing on the side of a street.
	\item A woman in a red jacket crossing the street
	\item a street light some people and a woman wearing a red jacket
	\item A blonde woman in a red coat crosses the street with her friend.
      \end{compactenumerate}
    &
      \begin{compactitemize}
        \item MLE: a woman walking down a street while holding an umbrella .
        \item PG-SPICE: a group of people walking down a street with a man on a street holding a traffic light and a traffic light on a city street with a city street
        \item MIXER-BCMR: a group of people walking down a street .
        \item MIXER-BCMR-A: a group of people walking down a street .
        \item PG-BCMR: a group of people walking down a city street .
        \item PG-SPIDEr: a group of people walking down a street with a traffic light .
      \end{compactitemize}
    \\ \hline


    \begin{minipage}{.2\textwidth}
      \includegraphics[width=\linewidth]{Figures/COCO_val2014_000000115146.jpg}
    \end{minipage}
    &
      \begin{compactenumerate}
        \item A group of people converse in an office setting.
        \item A group of people playing a game with remote controllers.
        \item Four young people have crowded into a small office.
        \item A group of people standing next to each other in a room.
        \item a group of people standing next to each other with some of them holding video game controllers
      \end{compactenumerate}
    &
      \begin{compactitemize}
        \item MLE: a group of people standing around a living room .
        \item PG-SPICE: a group of people in a room with a man in a chair holding a nintendo wii remote in a living room with a man in a chair holding a
        \item MIXER-BCMR: a group of people standing in a living room .
        \item MIXER-BCMR-A: a group of people standing in a living room playing a video game .
        \item PG-BCMR: a group of people standing in a room .
        \item PG-SPIDEr: a group of people playing a video game in a living room .
      \end{compactitemize}
    \\ \hline
    \begin{minipage}{.2\textwidth}
      \includegraphics[width=\linewidth]{Figures/COCO_val2014_000000099182.jpg}
    \end{minipage}
    &
      \begin{compactenumerate}
        \item A man looking through a book on top of a table. 
        \item A man sitting on a bed looking at a book 
        \item a man is flipping through a book on a bed 
        \item A man sitting on a bed flipping through pages of a book. 
        \item A man in a black jacket is flipping through a large book. 
      \end{compactenumerate}
    &
      \begin{compactitemize}
        \item MLE: a man sitting in front of a laptop computer .
        \item PG-SPICE: a man sitting in front of a book and a laptop on a table with a laptop computer on top of a table with a laptop computer on top of
        \item MIXER-BCMR: a man sitting in a chair with a book .		
        \item MIXER-BCMR-A: a man sitting at a table with a book .
        \item PG-BCMR: a man sitting in front of a book .
        \item PG-SPIDEr: a man sitting at a table with a book .
      \end{compactitemize}
    \\ \hline
  \end{tabular}
  \caption{Example captions from different models on COCO hold-out validation images. }
  \label{fig:examples}
\end{table*}

Table~\ref{fig:examples} shows some example captions generated by 6 different
methods: MLE, PG-SPICE, MIXER-BCMR, MIXER-BCMR-A, PG-BCMR, and PG-SPIDEr. 
We see that PG-SPICE tends to generate ungrammatical sentences, with a lot of repeated
phrases. This is because SPICE measures how well the scene graph induced by a
sentence matches the ground truth scene graph, but is relatively insensitive to
syntactic quality.  However, when we combine SPICE with CIDEr, we get much
better results.  We therefore ignore pure SPICE in the rest of this paper.

\begin{figure}
    \centering
    \includegraphics[width=\linewidth]{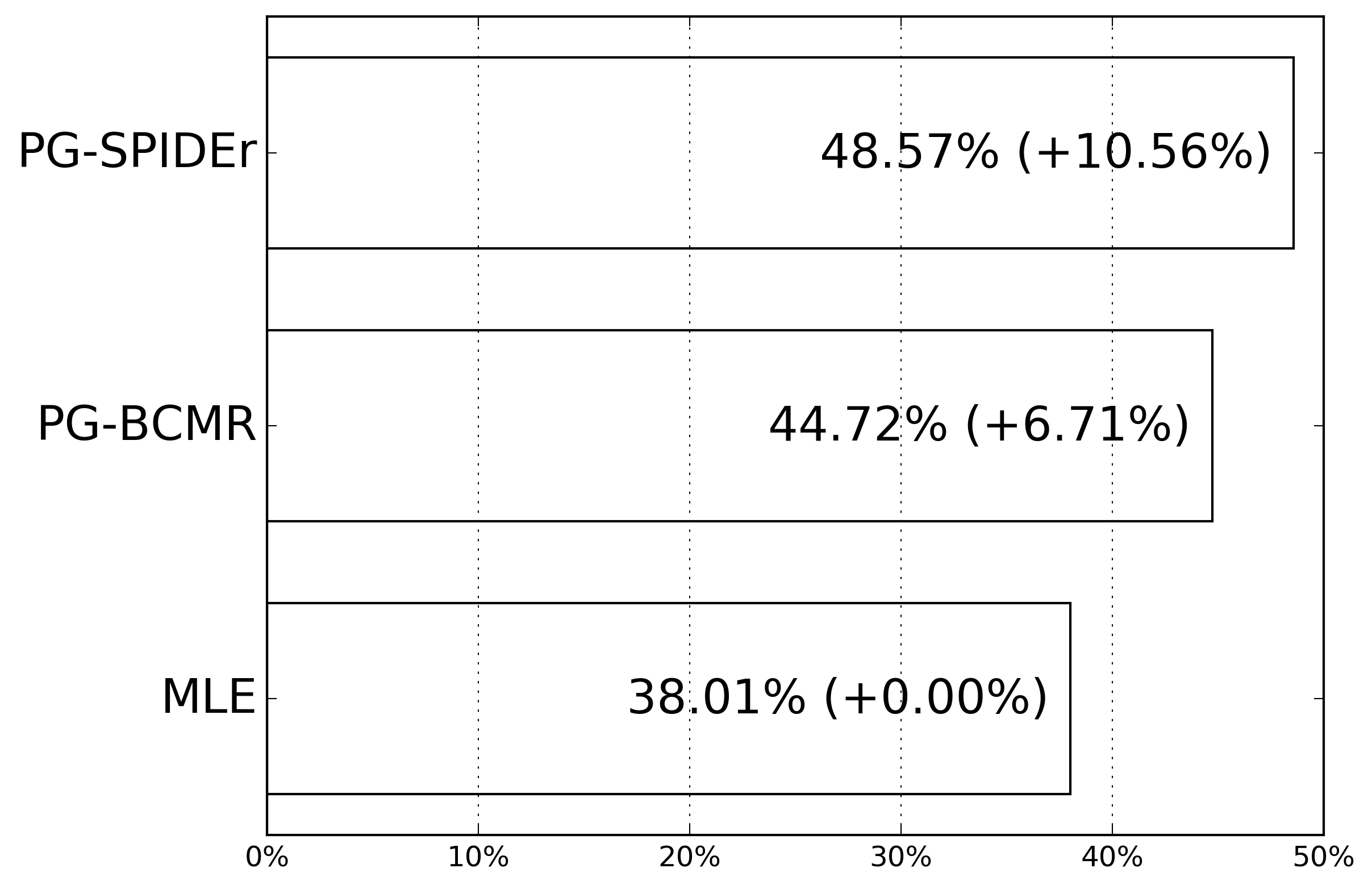}
    \caption{
 Results of human evaluation on 492 images randomly sampled from the
 COCO test set. We report the difference in percentage of
 ``not bad'' captions for each method compared to baseline 38\% of MLE 
 model.
    }
    \label{fig:cocoHuman}
\end{figure}

\begin{table}[ht]
\centering
\begin{tabular}{|c|c|c|c|c|c|}
\hline
p-value(X \textgreater Y) & PG-BCMR  & MLE  \\ \hline
PG-SPIDEr & \textbf{0.014} & \textbf{\textless 0.001}  \\ \hline
PG-BCMR & - & \textbf{0.003}  \\ \hline
\end{tabular}
\caption{p-values derived from a pairwise sign test applied
  to human ratings on 492 images from COCO test set. Statistically significant comparisons
  (at the 0.05 level) are shown in
  bold. X and Y correspond to rows and columns respectively.} 
\label{tab:p-value}
\end{table}

We also see that PG-SPIDEr tends to generate more reasonable
captions that PG-BCMR, even though it did worse on the COCO metrics.
Also, both methods seem to be better than
MLE. (See for example the third row in Table~\ref{fig:examples}.)

To quantify this, we turn to a user study.
In particular, we use a crowd sourcing platform,
using raters who
have prior  experience with evaluating image captioning and other
computer vision models.
We showed each image-caption pair to 3 different raters, and asked
them to evaluate it on a 4 point scale, depending on whether the
caption is ``bad'', ``okay'', ``good'' or ``excellent''.\footnote{
  The definitions of these terms, which we gave to raters,
  is as follows.
  Excellent: ``The caption correctly, specifically and completely
  describes the foreground/main objects/events/theme of the image.'' 
Good: ``The caption correctly and specifically describes most of the
foreground/main objects/events/theme of the image, but has minor
mistakes in some minor aspects.''
Okay: ``The caption correctly describes some of the foreground/main
objects/events/theme of the image, but is not specific to the image
and has minor mistakes in some minor aspects.''
Bad: ``The caption misses the foreground/main objects/events/theme of
the image or the caption contains obviously hallucinated
objects/activities/relationships.''
} %
We then take the majority vote to get the final rating.
If no majority is found, the rating is considered unknown,
and this image is excluded from the analysis. 

Since current captioning systems are far from perfect, 
our main goal is to develop a captioning system that does not
make ``embarrassing'' errors, we focus on measuring the fraction of
captions that are classified as ``not bad'',
which we interpret as the union of ``okay'', ``good'' and
``excellent''.

As ``quality control'',
we first evaluated 505 ground truth captions from the
COCO validation set. Humans said that 87\% of these captions were ``not bad''.  Some
of the 13\% of ground truth captions that were labeled ``bad'' do indeed contain errors\footnote{
  For example, some captions contain
  repeated words, e.g., ``this is an image image of a modern kitchen''.
  Others contain typos, e.g., ``a blue and white truck with pants in
  it's flat bed''.
  And some do not make semantic sense,
  e.g., ``A crowd of people parked near a double decker bus''.
}, %
due to the fact that COCO captions were generated by AMT workers who are not
perfect.  On the other hand, some captions seem reasonable to us, but did not
meet the strict quality criteria our raters were looking for.  In any case, 87\%
is an approximate upper bound on performance we can hope to achieve on the
COCO test set. 

We then randomly sampled 492 images from the test set (for which we do
not have access to the ground truth captions),
and generated captions from all of them using our 3 systems,
and sent them for human evaluation. 
Figure~\ref{fig:cocoHuman} shows the fraction of captions that are
``not bad'' compared to the MLE baseline of 38\%.
We draw the following conclusions:

\begin{compactitemize}

\item All methods are far below the human ceiling of 87\%.
    
\item All PG methods outperform MLE training by a significant margin (see
	Table~\ref{tab:p-value} for pairwise p-value analysis).  This is because
	the PG methods optimize metrics that are much more closely related to
	caption quality than the likelihood score.

\item PG-SPIDEr outperforms PG-BCMR by a 4\% margin, despite the fact that
	PG-BCMR outperforms PG-SPIDEr on all the COCO  metrics. This is
	because SPIDEr captures both fluency and semantic properties of the
	caption, both of which human raters are told to pay attention to,
	whereas BCMR is a more syntactic measure.

\end{compactitemize}

\subsection{Comparison with MIXER}
\label{sec:mixerResults}
\label{pg-mixer-comp}

We now compare our method with
\cite{Ranzato2015} in more detail.
When using MLE training, they get a BLEU-4 score of 0.278,
and when they  used MIXER to optimize BLEU-4,
they get 0.292, which is about 0.1 better.
(They do not report any other results besides BLEU-4.)
Note, however, that their numbers are not comparable to the numbers
in Table~\ref{fig:leaderboard},
since they are evaluated on 5000 images from the COCO validation set,
and not the official test server.

In order to do a fair comparison with our work (and other publications),
we reimplemented their algorithm.
When we used their train/test split\footnote{
  We thank Marc'Aurelio Ranzato for sharing their code and data split.
  }, 
and used our implementation
of MIXER to optimize BLEU-4, we were able to reproduce their BLEU-4 result.
We then used MIXER to optimize BCMR, using the standard train/test
split.
Once again, we see that the BLEU-4 score is about 0.1 better than MLE,
but the other metrics are sometimes worse,
whereas our PG method was significantly better than MLE across all metrics.

Upon digging into their open source code\footnote{
  \url{https://github.com/facebookresearch/MIXER}
},
we noticed that they used a different optimization algorithm:
we use Adam \cite{kingma2014adam}, whereas they used vanilla stochastic gradient descent
with a hard-coded learning rate decay.
When we reran MIXER-BCMR with Adam (a combination we call
MIXER-BCMR-A) with well tuned parameters, we saw significantly improved results,
which come closer to our best results 
obtained with PG-BCMR
in almost all metrics
(and even slightly beating us in Meteor).

\begin{figure}
	\centering
	\begin{subfigure}[b]{0.225\textwidth}
		\centering
		\includegraphics[width=1.0\textwidth]{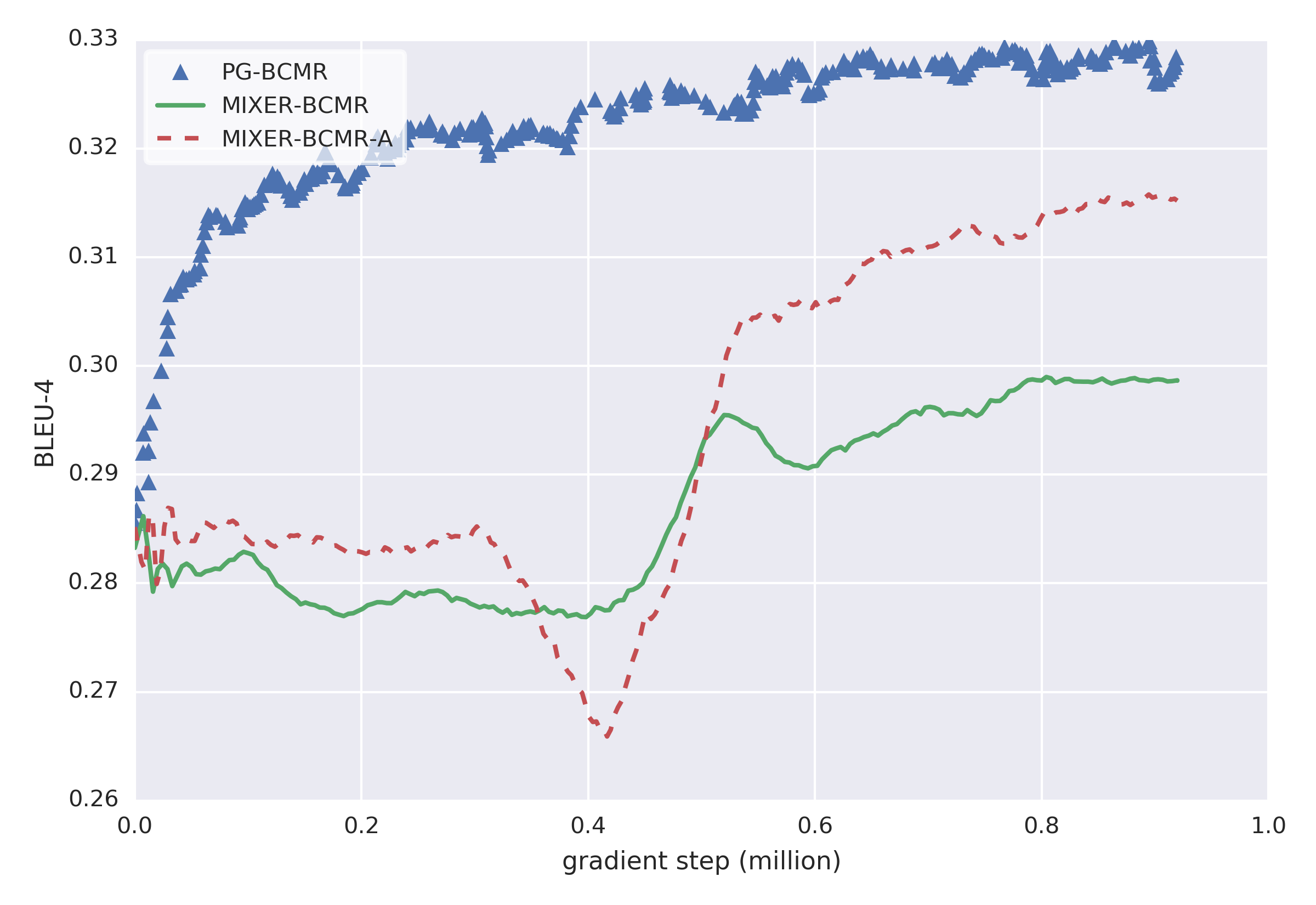}
		\caption[BLEU-4]%
		{{\small BLEU-4}}
		\label{fig:mixer-comp-bleu-4}
	\end{subfigure}
	\hfill
	\begin{subfigure}[b]{0.225\textwidth}
		\centering
		\includegraphics[width=1.0\textwidth]{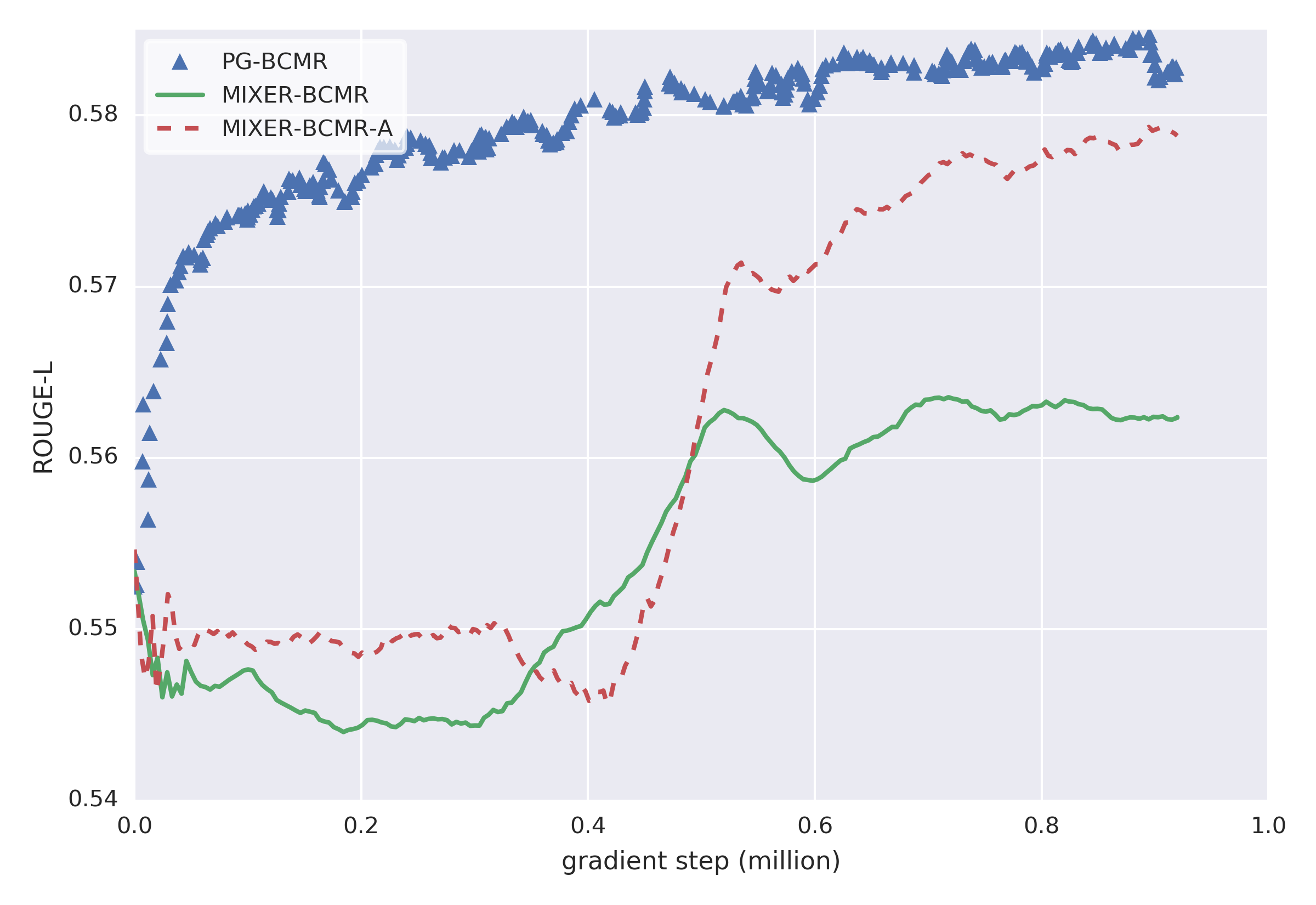}
		\caption[ROUGE-L]%
		{{\small ROUGE-L}}
		\label{fig:mixer-comp-rouge-l}
	\end{subfigure}
	\begin{subfigure}[b]{0.225\textwidth}
		\centering
		\includegraphics[width=1.0\textwidth]{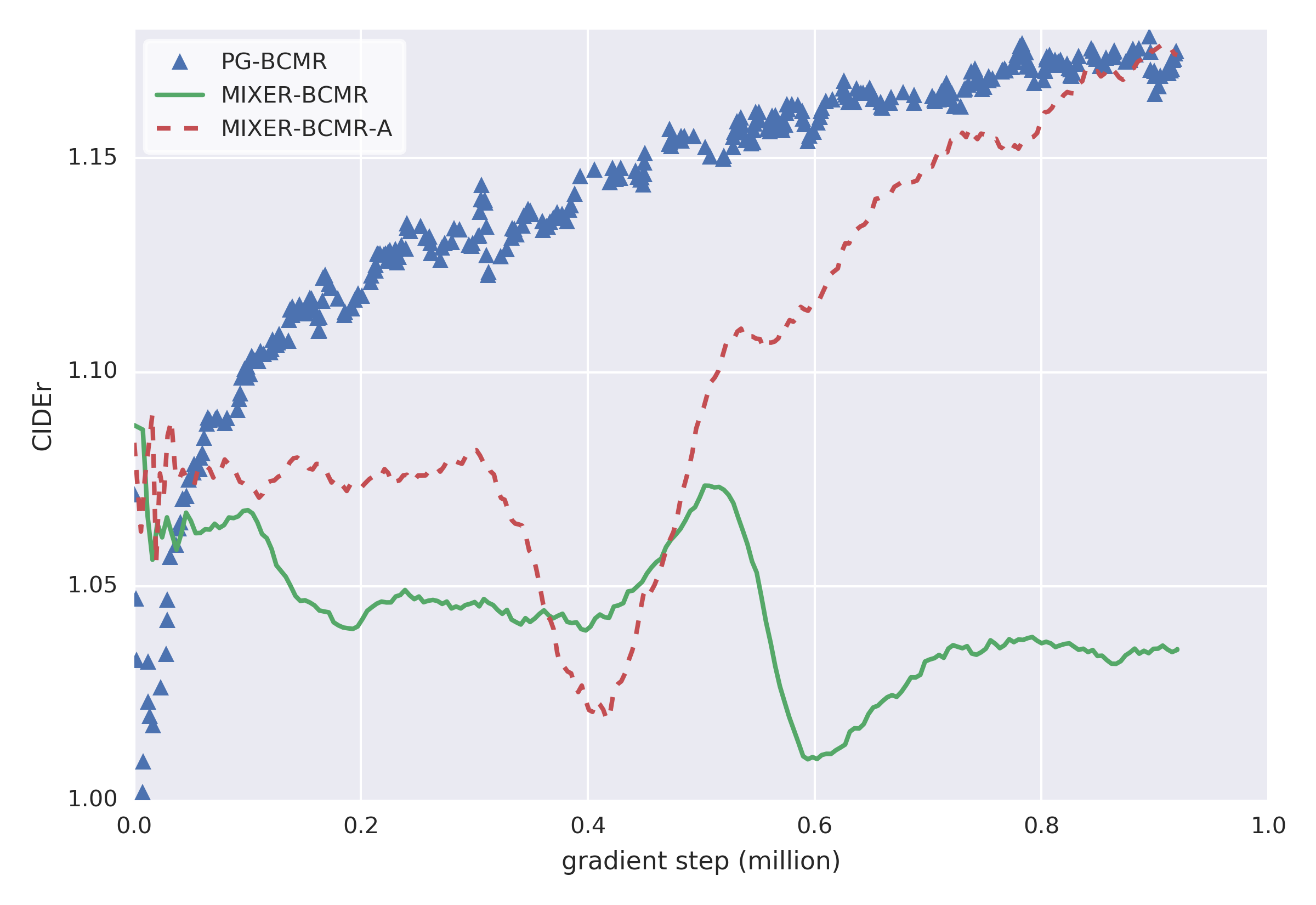}
		\caption[CIDEr]%
		{{\small CIDEr}}
		\label{fig:mixer-comp-rouge-l}
	\end{subfigure}
	\hfill
	\begin{subfigure}[b]{0.225\textwidth}
		\centering
		\includegraphics[width=1.0\textwidth]{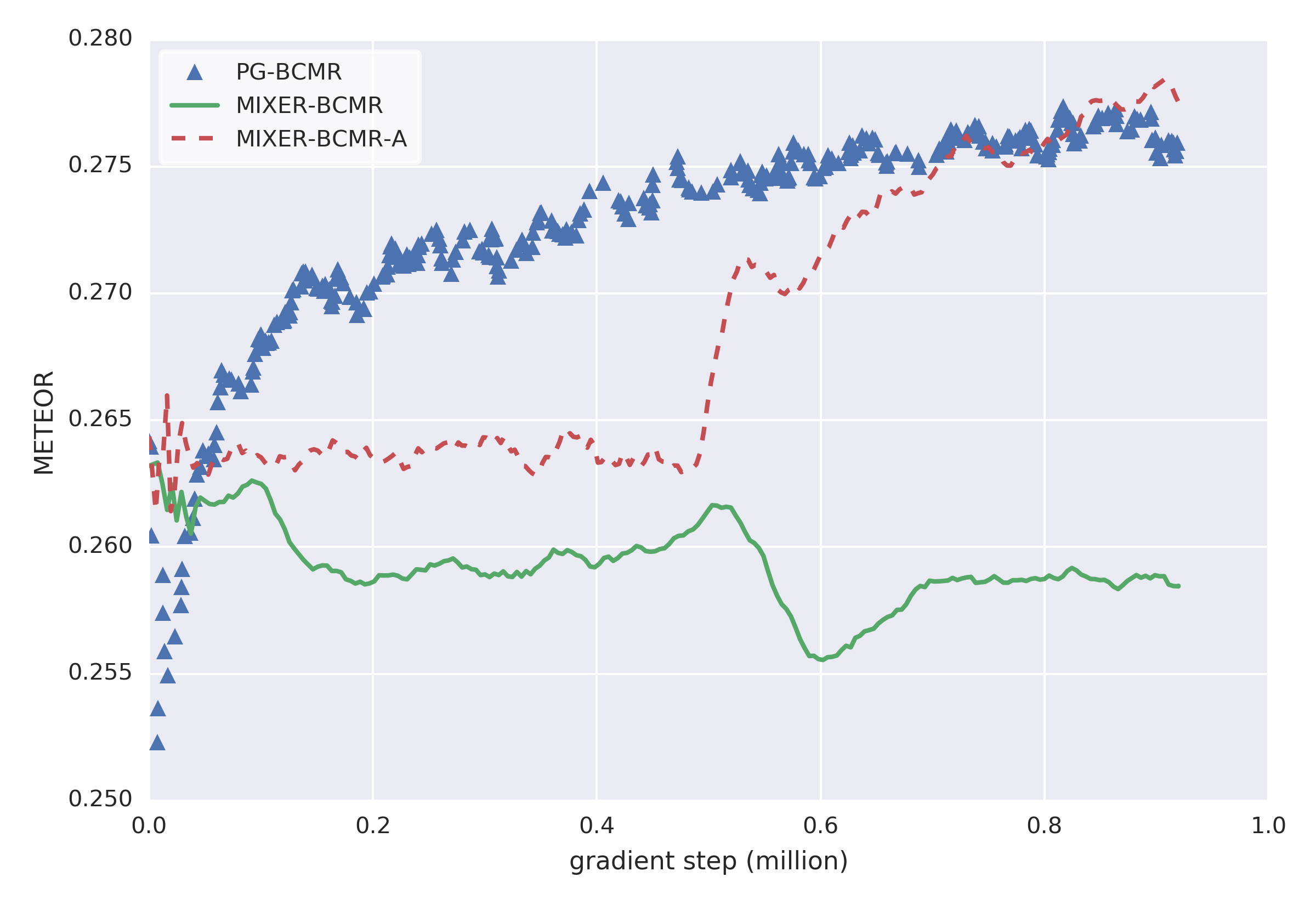}
		\caption[METEOR]%
		{{\small METEOR}}
		\label{fig:mixer-comp-spice}
	\end{subfigure}

	\caption{Performance of PG-BCMR (blue-triangle), MIXER-BCMR-A (red-dashed)
	and MIXER-BCMR (green-solid) on the validation set during  the first
	1 million gradient steps.}

 	\label{pg-mixer-comp-fig}
\end{figure}

However, the final results are not the entire story.
We have found that MIXER is much slower to converge,
and much less stable during training,
than our PG-Rollout method.
This is illustrated in
Figure~\ref{pg-mixer-comp-fig},
where we plot the BCMR metrics on the validation set
for three methods:
PG-BCMR (blue),
MIXER-BCMR-A (red),
MIXER-BCMR (green).
We show the best of 10 runs for MIXER,
and a single run for PG.
(Repeated runs of PG give similar performance.)

We see that MIXER-BCMR with vanilla SGD does poorly.  Using Adam helps a
lot, but performance is still very unstable; in particular, we had to try the
method 10 times, with different learning rates and gradient multipliers for the
baseline estimator, to get their best result shown in Figure~\ref{pg-mixer-comp-fig}. 

Note that the plateau in MIXER's performance curve for the first 500k steps is
because MIXER starts with just MLE training; no progress is made during this
time, since we initialize the models with a model that was already trained with
MLE.  At the 500k epoch, MIXER uses MLE for the first 6 words in the caption,
and REINFORCE for the rest.  At the 600k epoch, MIXER switches to REINFORCE for
the entire sequence.  Over time, the baseline estimator is learned, and this can
compensate for the inaccurate estimate of future rewards.  Hence eventually,
MIXER-BCMR-A can catch up with our PG-Rollout method.

\section{Conclusion}
\label{sec:concl}

In this paper, we have proposed a robust and efficient policy gradient method,
and successfully applied it to optimize a variety of  captioning metrics.
By optimizing the standard COCO metrics, we show that we can
achieve state of the art results, according to the leaderboard.
However, we also show that these metrics do not correlate well with human judgement.
We therefore also proposed a new metric, SPIDEr,
and show that optimizing this with our new algorithm
produces qualitatively superior results,
as judged by human raters.

\eat{
In this paper, we have proposed a robust and efficient policy gradient method
and successfully applied it to optimize all captioning metrics we tried, with
the goal of understanding what metrics correlate best with humans. We further
showed its success by optimizing towards COCO metrics, yielding competitive
methods to other state of the art systems. 

More importantly, our method enables us to draw two conclusions from this work.
First, a high score on a ``good'' metric should be a necessary and sufficient
condition of ``good'' captions, as perceived by humans. However, we show that SPICE does
not satisfy the latter and standard COCO metrics do not correlate well with
humans rating. Second, our proposed SPIDEr metric is clearly the preferred quality measure
by humans.
}

{\small
\bibliographystyle{ieee}
\bibliography{bib}
}

\newpage
\clearpage
\appendix
\section{Full derivation of policy gradient}
\label{sec:appendix_pg}
Given an image $\vx^n$ and the ground-truth captions $\vy^n$ associated
with $\vx^n$, the reward for a generated caption
$\bar{g}=g_{1:T}=\{g_1, g_2, ..., g_T\}$ is defined as
\be
R(\bar{g}|\vx^n,\vy^n) = \sum_{t=1}^{T} r(g_t|g_{1:(t-1)},\vx^n,\vy^n),
\label{eqn:reward}
\ee
where $r(g_t|g_{1:(t-1)},\vx^n,\vy^n)$ is the incremental reward for word
$g_t$, given previously generated words $g_{1:(t-1)}$ and $(\vx^n,\vy^n)$.
We define
\be
r(g_1|g_{1:0},\vx^n,\vy^n)=r(g_1|\vx^n,\vy^n)
\label{eqn:reward_edge_case}
\ee
for notation consistency.
Caption $\bar{g}$ is generated by sampling one word at each time step
from the given vocabulary $\calV$ using the conditional probability or policy
$\pi_{\theta}(\bar{g}|\vx^n)$ parameterized by $\theta$. Here $\theta$ is the set of
weights in the CNN-RNN neural network covered in section \ref{sec:model}.
The expected reward for the captions generated by following the policy
$\pi_{\theta}(\bar{g}|\vx^n)$ is
\be
V_{\theta}(\vx^n) = E_{\bar{g}}[R(\bar{g}|\vx^n,\vy^n)] = \sum_{\bar{g}}
                      \pi_{\theta}(\bar{g}|\vx^n) R(\bar{g}|\vx^n,\vy^n).
\label{eqn:expected_reward}
\ee
The goal is to maximize the objective function
\be
J(\theta) = \frac{1}{N} \sum_{n=1}^N V_{\theta}(\vx^n)
\label{eqn:objective}
\ee
where $N$ is the number of images in a given training data set.

In this paper we use the gradient-based methods to find the optimal
solution $\theta^*$.
In view of equation (\ref{eqn:expected_reward}), to compute the gradient
\be
\nabla_{\theta}J(\theta) = \frac{1}{N} \sum_{n=1}^N \nabla_{\theta} V_{\theta}(\vx^n),
\label{eqn:objective_gradient}
\ee
we just need to focus on the term
\begin{align}
& \nabla_{\theta} V_{\theta}(\vx^n)
  = \sum_{\bar{g}} \nabla_{\theta} \pi_{\theta}(\bar{g}|\vx^n) R(\bar{g}|\vx^n,\vy^n)
\label{eqn:reward_gradient}
 \\
  &= \sum_{\bar{g}} \pi_{\theta}(\bar{g}|\vx^n) \nabla_{\theta} \log(\pi_{\theta}(\bar{g}|\vx^n)) R(\bar{g}|\vx^n,\vy^n)
\nonumber \\
  &= E_{\bar{g}} [\nabla_{\theta} \log(\pi_{\theta}(\bar{g}|\vx^n)) R(\bar{g}|\vx^n,\vy^n)].
\label{eqn:reward_gradient_alg_1}
\end{align}
In view of the chain rule applied to the joint conditional probability
\be
  \pi_{\theta}(\bar{g}|\vx^n)
  = \prod_{t=1}^{T} \pi_{\theta}(g_t|g_{1:(t-1)},\vx^n),
  \label{eqn:chain_rule}
\ee
we have
\be
\nabla_{\theta} \log(\pi_{\theta}(\bar{g}|\vx^n)) =
   \sum_{t=1}^{T} \nabla_{\theta} \log(\pi_{\theta}(g_t|g_{1:(t-1)},\vx^n)).
\label{eqn:accumulated_log_pi}
\ee
Similar to equation (\ref{eqn:reward_edge_case}), we define
\be
\pi_{\theta}(g_1|g_{1:0},\vx^n)=\pi_{\theta}(g_1|\vx^n).
\label{eqn:chain_rule_edge_case_1}
\ee
Equations (\ref{eqn:reward_gradient_alg_1}) and (\ref{eqn:accumulated_log_pi})
suggest a straight forward sequence-level Monte Carlo sampling algorithm.
But we will show next that this can be further improved.

Taking the derivative on both sides of the equation (\ref{eqn:chain_rule})
leads to
\begin{align}
&\nabla_{\theta} \pi_{\theta}(\bar{g}|\vx^n)
= \nabla_{\theta} \pi_{\theta}(g_1|\vx^n) \prod_{t=2}^{T} \pi_{\theta}(g_t|g_{1:(t-1)},\vx^n)
\nonumber \\
&+ \pi_{\theta}(g_1|\vx^n) \nabla_{\theta} \pi_{\theta}(g_2|g_1,\vx^n) \prod_{t=3}^{T} \pi_{\theta}(g_t|g_{1:(t-1)},\vx^n)
\nonumber \\
&+ \ldots
\nonumber \\
&+ \prod_{t=1}^{T-1} \pi_{\theta}(g_t|g_{1:(t-1)},\vx^n) \nabla_{\theta} \pi_{\theta}(g_T|g_{1:(T-1)},\vx^n) 
\nonumber \\
&= \sum_{t=1}^{T} \pi_{\theta}(g_{1:(t-1)}|\vx^n)
                \nabla_{\theta} \pi_{\theta}(g_t|g_{1:(t-1)},\vx^n)
\nonumber \\
&  \times
               \pi_{\theta}(g_{t+1:T}|g_{1:t},\vx^n),
\label{eqn:chain_rule_derivative}
\end{align}
where the first equality is due to simple calculus rules,
and the second equality is due to the chain rule in equation
(\ref{eqn:chain_rule}). Here we define
\begin{align}
&\pi_{\theta}(g_{1:0}|\vx^n)=1.
\label{eqn:chain_rule_edge_case_2}
\\
&\pi_{\theta}(g_{(T+1):T}|g_{1:T},\vx^n)=1.
\label{eqn:chain_rule_edge_case_3}
\end{align}
for notation consistency in equation (\ref{eqn:chain_rule_derivative}).

To simplify the notation and unless confusion arises,
we will drop the conditional terms $(\cdot|\vx^n,\vy^n)$
and $(\cdot|\vx^n)$ in the following derivation.

Substituting equation (\ref{eqn:chain_rule_derivative}) into equation (\ref{eqn:reward_gradient})
and changing the order of summation, we obtain
\begin{align}
&  \nabla_{\theta} V_{\theta}(\vx^n)
  =
  \sum_{t=1}^{T} 
  \sum_{g_{1:(t-1)}} \pi_{\theta}(g_{1:(t-1)}) 
  \sum_{g_t} \nabla_{\theta} \pi_{\theta}(g_t|g_{1:(t-1)})
\nonumber \\
&  \times
\sum_{g_{(t+1):T}} \pi_{\theta}(g_{t+1:T}|g_{1:t}) R(\bar{g})
\nonumber \\
& =
  \sum_{t=1}^{T} 
  E_{g_{1:(t-1)}}[
    E_{g_{t}}[\nabla_{\theta} \log(\pi_{\theta}(g_t|g_{1:(t-1)}))
      E_{g_{(t+1):T}} [R(\bar{g})]
    ]
  ].
\label{eqn:reward_derivative_expanded}
\end{align}
Note that 
In view of equation (\ref{eqn:reward}), we can partition the reward
$R(\bar{g})$ in equation (\ref{eqn:reward_derivative_expanded}) into two parts 
\be
  R(\bar{g}) = R(g_{1:(t-1)}) + R(g_{1:(t-1)},g_{t:T}),
\label{eqn:partitioned_reward}
\ee
where
\be
R(g_{1:(t-1)},g_{t:T}) = \sum_{i=t}^{T} r(g_i|g_{1:(i-1)}).
\label{eqn:reward_part_2}
\ee
Here we want to show that the contribution of $R(g_{1:(t-1)})$ to the summation
in equation (\ref{eqn:reward_derivative_expanded}) is zero.
Specifically, the inner summation for $R(g_{1:(t-1)})$ becomes
\begin{align}
&  E_{g_t} [\nabla_{\theta} \log(\pi_{\theta}(g_t|g_{1:(t-1)}))
   E_{g_{(t+1):T}} [R(g_{1:(t-1)})]]
\nonumber \\
&=
R(g_{1:(t-1)}) E_{g_t} [\nabla_{\theta} \log(\pi_{\theta}(g_t|g_{1:(t-1)})) E_{g_{(t+1):T}} [1]]
\nonumber \\
&=
R(g_{1:(t-1)}) \sum_{g_t} [\nabla_{\theta} \pi_{\theta}(g_t|g_{1:(t-1)})]
\nonumber \\
&=
 R(g_{1:(t-1)})  \nabla_{\theta} E_{g_t}[1] = 0,
\label{eqn:reward_derivative_part_1}
\end{align}
where the first equality is due to the fact that $R(g_{1:(t-1)})$ is independent of
$g_t$ and $g_{(t+1):T}$, the second equality is due to the definition of expectation,
and the third equality is due to the exchange of sum and derivative.
The contribution of $R(g_{1:(t-1)},g_{t:T})$ to the summation in equation
(\ref{eqn:reward_derivative_expanded}) can be simplified as well.
Specifically, the inner most summation becomes
\begin{align}
& Q_{\theta}(g_{1:(t-1)},g_{t}) = E_{g_{(t+1):T}} [R(g_{1:(t-1)},g_{t:T})]
\nonumber \\
& = E_{g_{(t+1):T}} [r(g_{t}|g_{1:(t-1)}) + R(g_{1:t},g_{(t+1):T})]
\nonumber \\
& = r(g_{t}|g_{1:(t-1)}) E_{g_{(t+1):T}}[1] + E_{g_{(t+1):T}} [R(g_{1:t},g_{(t+1):T})]
\nonumber \\
& = r(g_{t}|g_{1:(t-1)}) + E_{g_{(t+1):T}} [R(g_{1:t},g_{(t+1):T})],
\label{eqn:reward_derivative_part_2}
\end{align}
where the second equality is due to the definition in equation (\ref{eqn:reward_part_2}),
and the third equality is due to the fact that $r(g_{t}|g_{1:(t-1)})$ is
independent of $g_{(t+1):T}$. Hence equation (\ref{eqn:reward_derivative_expanded})
becomes
\begin{align}
&  \nabla_{\theta} V_{\theta}(\vx^n)
  =
  \sum_{t=1}^{T} 
  E_{g_{1:(t-1)}} [
    E_{g_t} [\nabla_{\theta} \log(\pi_{\theta}(g_t|g_{1:(t-1)}))
\nonumber \\
&  \times      
      Q_{\theta}(g_{1:(t-1)},g_{t})
    ]
  ]
\label{eqn:reward_gradient_alg_2}  \\
& =
  \sum_{t=1}^{T} 
  E_{g_{1:T}} [
    E_{g_t} [\nabla_{\theta}\log(\pi_{\theta}(g_t|g_{1:(t-1)}))
      Q_{\theta}(g_{1:(t-1)},g_{t})
    ]
  ]
\nonumber \\
& =
  E_{g_{1:T}} [
  \sum_{t=1}^{T} 
    \sum_{g_t \in \calV} [\nabla_{\theta} \pi_{\theta}(g_t|g_{1:(t-1)})
      Q_{\theta}(g_{1:(t-1)},g_{t})
    ]
  ],
\label{eqn:reward_gradient_full}
\end{align}
where the second equality is due to the fact that the term
$E_{g_t} [\nabla_{\theta}\log(\pi_{\theta}(g_t|g_{1:(t-1)}))Q_{\theta}(g_{1:(t-1)},g_{t})]$ 
is independent of $g_{t:T}$.
Equation (\ref{eqn:reward_gradient_full}) is also shown in \cite{Bahdanau2016}.


\section{Variance reduction}
\label{sec:appendix_variance_reduction}
The Monte Carlo sampling used to compute the expectation in equations
(\ref{eqn:reward_derivative_part_2})
and (\ref{eqn:reward_gradient_alg_2}) typically suffers from high variance
problem due to very high dimensionality of the sample space.
For example, the vocabulary of the COCO data set is about nine thousand.
If we set caption length $T=30$, then the discrete sample space has
$9000^{30}$ grid points. 
One commonly used solution is to subtract a constant from $Q_{\theta}(g_{1:(t-1)},g_{t})$ in
equation (\ref{eqn:reward_gradient_alg_2}). The inner summation in equation
(\ref{eqn:reward_gradient_alg_2}) now becomes
\be
E_{g_t} [\nabla_{\theta}\log(\pi_{\theta}(g_t|g_{1:(t-1)})) (Q_{\theta}(g_{1:(t-1)},g_{t}) - B)].
\label{eqn:reward_gradient_debiased}
\ee
Since the second term is
\be
E_{g_t} [\nabla_{\theta}\log(\pi_{\theta}(g_t|g_{1:(t-1)})) B]
= \nabla_{\theta} E_{g_t}[B] = 0,
\label{eqn:reward_gradient_bias}
\ee
this manipulation does not theorectically change the gradient in equation
(\ref{eqn:reward_gradient_alg_2}).
Prior work in \cite{zaremba2015reinforcement, Williams1992} show that
\be
  B = E_{g_t} [Q_{\theta}(g_{1:(t-1)},g_{t})]
\label{eqn:expected_baseline_bias}
\ee
is effective in reducing the variance due to small sample size.

\end{document}